\documentclass[10pt,twocolumn,letterpaper]{article}

\usepackage{iccv}
\usepackage{times}
\usepackage{epsfig}
\usepackage{graphicx}
\usepackage{amsmath}
\usepackage{amssymb}
\usepackage{multirow}
\usepackage{moresize}

\usepackage[pagebackref=true,breaklinks=true,letterpaper=true,colorlinks,bookmarks=false]{hyperref}

\iccvfinalcopy 

\def\iccvPaperID{1884}

\ificcvfinal\fi

\newcommand{\fig}[1]{Fig.~\ref{fig:#1}}
\newcommand{\tab}[1]{Table~\ref{tab:#1}}
\newcommand{\secc}[1]{Section~\ref{sec:#1}}

\def\etal{{\textit{et~al.~}}}
\def\aal{\&{\textit{al.}~}}
\def\ie{{\textit{i.e.~}}}
\def\eg{{\textit{e.g.~}}}

\newcommand\enum[1]{({\it #1})}

\begin{document}

\title{
    Predicting Depth, Surface Normals and Semantic Labels \\
    with a Common Multi-Scale Convolutional Architecture
}

\author{
David Eigen$^1$ \quad Rob Fergus$^{1,2}$\\
$^1$ Dept. of Computer Science, Courant Institute, New York University\\
$^2$ Facebook AI Research\\
{\tt\small \{deigen,fergus\}@cs.nyu.edu}
\vspace{-4mm}
}

\maketitle

\begin{abstract}

In this paper we address three different computer vision tasks using a single
multiscale convolutional network architecture:  depth prediction, surface
normal estimation, and semantic labeling.  The network that we develop is able
to adapt naturally to each task using only small modifications, regressing from
the input image to the output map directly.  Our method progressively refines
predictions using a sequence of scales, and captures many image details without
any superpixels or low-level segmentation.  We achieve state-of-the-art
performance on benchmarks for all three tasks.

\end{abstract}

\vspace{-5mm}
\section{Introduction}
\vspace{-1mm}

Scene understanding is a central problem in vision that has many
different aspects. These include semantic labels describing the
identity of different scene portions; surface normals or depth estimates
describing the physical geometry; instance labels of the extent of individual
objects; and affordances capturing possible interactions of people with the
environment.  Many of these are often represented with a pixel-map containing a
value or label for each pixel, \eg a map containing the semantic label of the
object visible at each pixel, or the vector coordinates of the surface normal
orientation.

In this paper, we address three of these tasks, depth prediction, surface
normal estimation and semantic segmentation --- all using a single common
architecture.  Our multiscale approach generates pixel-maps directly from an
input image, without the need for low-level superpixels or contours, and is
able to align to many image details using a series of convolutional network
stacks applied at increasing resolution.  At test time, all three outputs can
be generated in real time ($\sim$30Hz).  We achieve state-of-the art results on
all three tasks we investigate, demonstrating our model's versatility. 

There are several advantages in developing a general model for pixel-map
regression.  First, applications to new tasks may be quickly developed, with
much of the new work lying in defining an appropriate training set and loss
function; in this light, our work is a step towards building off-the-shelf
regressor models that can be used for many applications.  In addition, use of a
single architecture helps simplify the implementation of systems
that require multiple modalities, e.g. robotics or augmented reality, which in
turn can help enable research progress in these areas.  Lastly, in the case of
depth and normals, much of the computation can be shared between modalities,
making the system more efficient.

\vspace{-1mm}
\section{Related Work}
\vspace{-1mm}

Convolutional networks have been applied with great success for object
classification and detection
\cite{Kriz12,rcnn,overfeat,vggimagenet,googlelenet}.  Most such systems
classify either a single object label for an entire input window, or bounding
boxes for a few objects in each scene.  However, ConvNets have recently been
applied to a variety of other tasks, including pose estimation
\cite{tompson2014joint,osadchy2006synergistic}, stereo depth
\cite{zbontar14,Memisevic11}, and instance segmentation \cite{gupta14}.  Most
of these systems use ConvNets to find only local features, or generate
descriptors of discrete proposal regions; by contrast, our network uses both
local and global views to predict a variety of output types.  In addition,
while each of these methods tackle just one or two tasks at most, we are able to
apply our network to three disparate tasks.

Our method builds upon the approach taken by Eigen \etal \cite{depth}, who apply
two convolutional networks in stages for single-image depth map prediction.  We
develop a more general network that uses a sequence of three scales to generate
features and refine predictions to higher resolution, which we apply to
multiple tasks, including surface normals estimation and per-pixel semantic
labeling.  Moreover, we improve performance in depth prediction as well,
illustrating how our enhancements help improve all tasks.

Single-image surface normal estimation has been addressed by Fouhey \etal
\cite{fouhey13,fouhey14}, Ladicky \etal \cite{ladicky14normals}, Barron and Malik \cite{barron15,barron13}, and most
recently by Wang \etal \cite{wang14}, the latter in work concurrent with ours.
Fouhey \etal match to discriminative local templates \cite{fouhey13} followed
by a global optimization on a grid drawn from vanishing point rays
\cite{fouhey14}, while Ladicky \etal learn a regression from over-segmented
regions to a discrete set of normals and mixture coefficients.  
Barron and Malik
\cite{barron15,barron13} infer normals from RGB-D inputs using a set of handcrafted priors, along with illumination and reflectance.
From RGB inputs, Wang \etal
\cite{wang14} use convolutional networks to combine normals estimates from
local and global scales, while also employing cues from room layout, edge
labels and vanishing points.  
Importantly, we achieve as good or superior
results with a more general multiscale architecture that can naturally be used
to perform many different tasks.

Prior work on semantic segmentation includes many different approaches, both
using RGB-only data \cite{tighe2013finding,cpmc,farabet2012scene} as well as
RGB-D
\cite{Silberman12,ren2012rgbd,mueller,couprie13,khan14,hermans14,gupta13}.
Most of these use local features to classify over-segmented regions, followed
by a global consistency optimization such as a CRF.  By comparison, our method
takes an essentially inverted approach:  We make a consistent global prediction
first, then follow it with iterative local refinements.  In so doing, the local
networks are made aware of their place within the global scene, and can can
use this information in their refined predictions.

Gupta \etal \cite{gupta13,gupta14} create semantic segmentations first by
generating contours, then classifying regions using either hand-generated
features and SVM \cite{gupta13}, or a convolutional network for object
detection \cite{gupta14}.  Notably, \cite{gupta13} also performs amodal
completion, which transfers labels between disparate regions of the image 
by comparing planes from the depth.

Most related to our method in semantic segmentation are other approaches using
convolutional networks.  Farabet \etal \cite{farabet2012scene} and Couprie
\etal \cite{couprie13} each use a convolutional network applied to multiple
scales in parallel generate features, then aggregate predictions using
superpixels.  Our method differs in several important ways.  First, our model
has a large, full-image field of view at the coarsest scale; as we demonstrate,
this is of critical importance, particularly for depth and normals tasks.  In
addition, we do not use superpixels or post-process smoothing --- instead, our
network produces fairly smooth outputs on its own, allowing us to take a simple
pixel-wise maximum.

Pinheiro \etal \cite{pinheiro14} use a recurrent convolutional network in which
each iteration incorporates progressively more context, by combining a more
coarsely-sampled image input along with the local prediction from the previous
iteration.  This direction is precisely the reverse of our approach, which
makes a global prediction first, then iteratively refines it.  In addition,
whereas they apply the same network parameters at all scales, we learn distinct
networks that can specialize in the edits appropriate to their stage.

Most recently, in concurrent work, Long \etal \cite{long14} adapt the recent
VGG ImageNet model \cite{vggimagenet} to semantic segmentation by applying 1x1
convolutional label classifiers at feature maps from different layers,
corresponding to different scales, and averaging the outputs.  By contrast, we
apply networks for different scales in series, which allows them to make more
complex edits and refinements, starting from the full image field of view.
Thus our architecture easily adapts to many tasks, whereas by considering
relatively smaller context and summing predictions, theirs is specific to
semantic labeling.

\begin{figure}
\centering
\includegraphics[width=\linewidth]{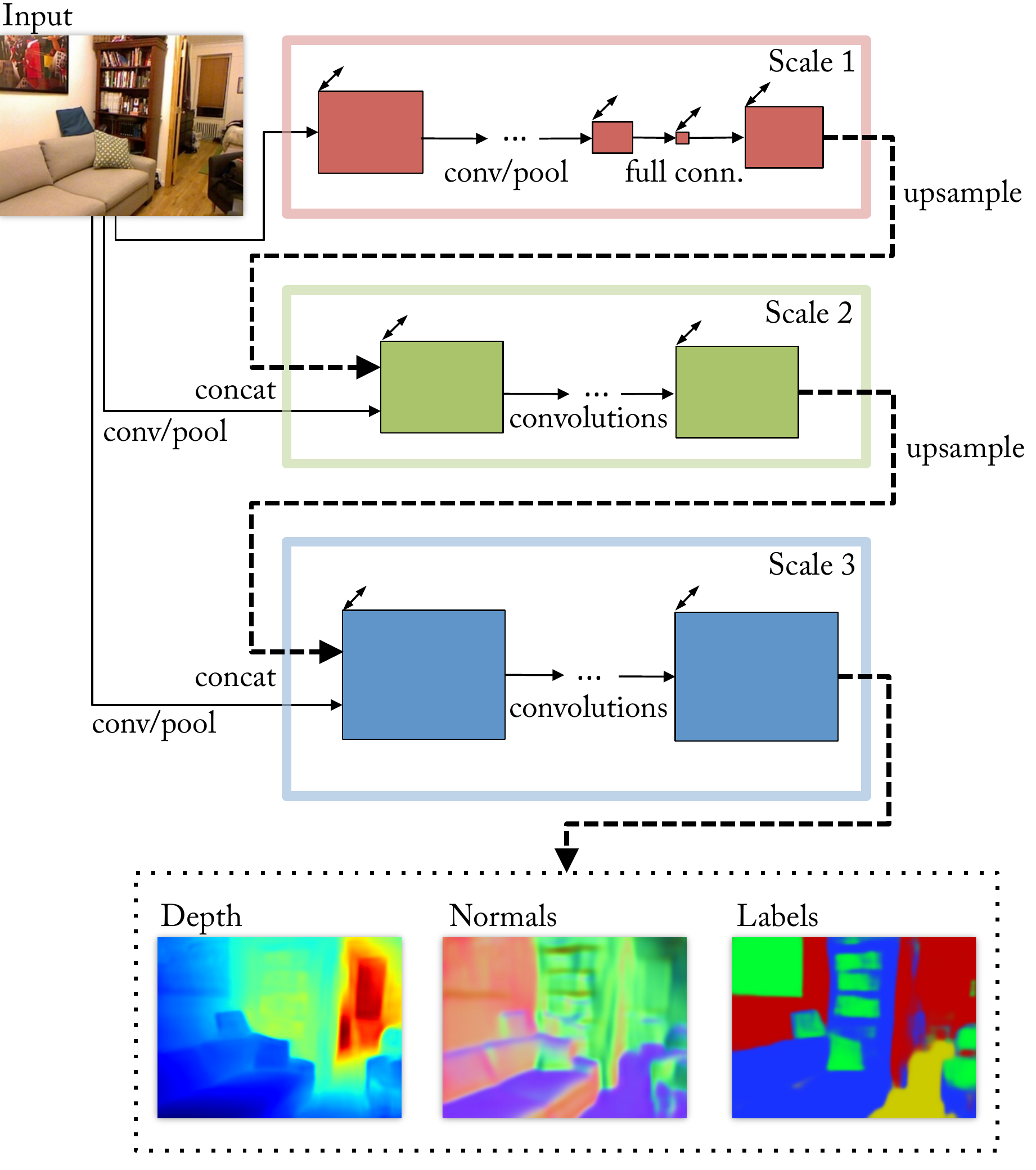}
\vspace{0pt}

{
\ssmall
\setlength{\tabcolsep}{2pt}
\begin{tabular}{|l|l|ccccccc|c|}
\hline
 & Layer & 1.1 & 1.2 & 1.3 & 1.4 & 1.5 & 1.6 & 1.7 & upsamp \\
 \hline
\multirow{3}{*}{Scale 1}
 & Size   & 37x27 & 18x13 & 18x13 & 18x13  & 8x6 & 1x1 & 19x14 & 74x55 \\
\multirow{3}{*}{(AlexNet)}
 & \#convs & 1 & 1 & 1 & 1 & 1 & -- & -- & -- \\
 & \#chan & 96 & 256 & 384 & 384 & 256 & 4096 & 64 & 64 \\
 & ker. sz & 11x11      & 5x5   & 3x3   & 3x3   & 3x3 & -- & -- & -- \\
 & Ratio  & /8 & /16 & /16 & /16  & /32   & -- & /16 & /4\\
 & l.rate & 0.001 & 0.001 & 0.001 & 0.001 & 0.001 & \multicolumn{2}{c|}{see text} & \\
\hline
 & Layer & 1.1 & 1.2 & 1.3 & 1.4 & 1.5 & 1.6 & 1.7 & upsamp \\
 \hline
\multirow{3}{*}{Scale 1}
 & Size    & 150x112  & 75x56 & 37x28 & 18x14 & 9x7 & 1x1  & 19x14 & 74x55  \\
\multirow{3}{*}{(VGG)}
 & \#convs & 2  & 2     & 3     & 3     & 3   & --   & --    & --     \\
 & \#chan  & 64       & 128   & 256   & 512   & 512 & 4096 & 64    & 64     \\
 & ker. sz  & 3x3      & 3x3   & 3x3   & 3x3   & 3x3 & -- & -- & -- \\
 & Ratio   & /2       & /4    & /8    & /16   & /32 & --   & /16   & /4\\
 & l.rate  & 0.001 & 0.001 & 0.001 & 0.001 & 0.001 & \multicolumn{2}{c|}{see text} & \\
\hline
\hline
\multirow{5}{*}{Scale 2}
 & Layer & 2.1 & 2.2 & 2.3 & 2.4 & 2.5  &  &  & upsamp \\
 \hline
 & Size  & 74x55 & 74x55 & 74x55  & 74x55  &74x55    && &   147x109 \\
 & \#chan & 96+64 & 64 & 64  & 64 & $C$ & & & $C$ \\
 & ker. sz  & 9x9 & 5x5   & 5x5   & 5x5   & 5x5 & & & -- \\
 & Ratio & /4 &  /4 & /4 &  /4 &  /4 &   &  & /2 \\
 & l.rate   & 0.001  & 0.01 & 0.01 & 0.01 & 0.001 & &  &  \\
\hline
\hline
\multirow{5}{*}{Scale 3}
 & Layer & 3.1 & 3.2 & 3.3 & 3.4 &  &  & & final\\
 \hline
 & Size  & 147x109 & 147x109  & 147x109  & 147x109  & & &  & 147x109  \\
 & \#chan & 96+$C$ & 64 & 64 & $C$ & & & & $C$ \\
 & ker. sz  & 9x9 & 5x5   & 5x5   & 5x5   &     & & & -- \\
 & Ratio & /2 & /2  & /2  & /2 &  &  &  &  /2 \\
 & l.rate   & 0.001  & 0.01 & 0.01 & 0.001 &  &  &  & \\
\hline
\end{tabular}
}
\caption{Model architecture.  $C$ is the number of output channels in the final prediction, which depends on the task.  The input to the network is 320x240.}
\label{fig:arch1}
\vspace{-5mm}
\end{figure}

\vspace{-1mm}
\section{Model Architecture}
\vspace{-1mm}

Our model is a multi-scale deep network that first
predicts a coarse global output based on the entire image area, then refines it
using finer-scale local networks.  This scheme is illustrated in \fig{arch1}.
While our model was initially based upon the architecture proposed by \cite{depth},
it offers several architectural improvements.
First, we make the model deeper (more convolutional layers). Second, we
add a third scale at higher resolution, bringing the final output resolution up
to half the input, or $147\times109$ for NYUDepth. Third, instead of passing
output \emph{predictions} from scale 1 to scale 2, we pass multichannel
\emph{feature maps}; in so doing, we found we could also train the first two
scales of the network jointly from the start, somewhat simplifying the training
procedure and yielding performance gains.

{\bf Scale 1:  Full-Image View}~The first scale in the network predicts a coarse but
spatially-varying set of features for the entire image area, based on a large, full-image
field of view, which we accomplish this through the use of two fully-connected
layers.  The output of the last full
layer is reshaped to 1/16-scale in its spatial dimensions by
$64$ features, then upsampled by a factor of 4 to 1/4-scale.
Note since the feature upsampling is linear, this corresponds to a
decomposition of a big fully connected layer from layer 1.6
to the larger $74\times55$ map;  since such a matrix would be
prohibitively large and only capable of
producing a blurry output given the more constrained
input features, we constrain the resolution
and upsample.  Note, however, that the 1/16-scale output is still large
enough to capture considerable spatial variation, and in fact is twice as large
as the 1/32-scale final convolutional features of the coarse stack.

Since the top
layers are fully connected, each spatial location in the output connects to
the all the image features, incorporating a very large field of view.
This stands in contrast to the multiscale approach of
\cite{couprie13,farabet2012scene}, who produce maps where the field of
view of each output location is a more local region centered on the output pixel.  
This full-view connection is especially important for depth and normals tasks,
as we investigate in \secc{scales}.

As shown in \fig{arch1}, we trained two different sizes of our model:  One
where this scale is based on an ImageNet-trained AlexNet \cite{Kriz12}, and one
where it is initialized using the Oxford VGG network \cite{vggimagenet}.  We
report differences in performance between the models on all tasks, to measure
the impact of model size in each.

{\bf Scale 2:  Predictions}~ The job of the second scale is to produce predictions at a mid-level
resolution, by incorporating a more detailed but narrower view of the
image along with the full-image information supplied by the coarse network.
We accomplish this by concatenating the feature maps of the coarse network with
those from a single layer of convolution and pooling, performed at finer stride
(see \fig{arch1}).
The output of the second scale is a 55x74 prediction (for NYUDepth), with
the number of channels depending on the task.
We train Scales 1 and 2 of the model together jointly, using SGD on the losses
described in \secc{tasks}.

{\bf Scale 3:  Higher Resolution}~
The final scale of our model refines the predictions to
higher resolution.  We concatenate the
Scale-2 outputs with feature maps generated from the original input at yet
finer stride, thus incorporating a more detailed view of the image.  The
further refinement aligns the output to higher-resolution details,
producing spatially coherent yet quite detailed outputs.  The final
output resolution is half the network input.

\vspace{-2mm}
\section{Tasks}
\vspace{-2mm}
\label{sec:tasks}

We apply this same architecture structure to each of the three tasks we
investigate:  depths, normals and semantic labeling.  Each makes use of a
different loss function and target data defining the task.

\vspace{-1mm}
\subsection{Depth}
\vspace{-2mm}

For depth prediction, we use a loss function comparing the predicted
and ground-truth log depth maps $D$ and $D^*$. Letting $d = D - D^*$ be
their difference, we set the loss to
\vspace{-3mm}
\begin{multline}
L_{depth}(D, D^*) ~ = ~ \frac1n \sum_i d_i^2 - \frac{1}{2n^2} \left(\sum_i d_i \right)^2 \\
           +  \frac1n\sum_i [(\nabla_x d_i)^2 + (\nabla_y d_i)^2]
\label{eqn:depthloss}
\end{multline}
where the sums are over valid pixels $i$ and $n$ is the number of
valid pixels (we mask out pixels where the
  ground truth is missing).  Here, $\nabla_x d_i$ and $\nabla_y d_i$ are the horizontal and
vertical image gradients of the difference.

Our loss is similar to that of \cite{depth}, who also use the $l_2$ and
scale-invariant difference terms in the first line.  However, we also include a
first-order matching term $(\nabla_x d_i)^2 + (\nabla_y d_i)^2$, which compares
image gradients of the prediction with the ground truth.  This encourages
predictions to have not only close-by values, but also similar local structure.
We found it indeed produces outputs that better follow depth gradients, with no
degradation in measured $l_2$ performance.

\vspace{-1mm}
\subsection{Surface Normals}
\vspace{-2mm}

To predict surface normals, we change the output from one channel to three, and
predict the $x$, $y$ and $z$ components of the normal at each pixel.  We also
normalize the vector at each pixel to unit $l_2$ norm, and backpropagate through this
normalization.  We then employ a simple elementwise loss comparing the
predicted normal at each pixel to the ground truth, using a dot product:
\vspace{-1mm}
\begin{equation}
L_{normals}(N, N^*) = -\frac1n \sum_i N_i \cdot N_i^*
                    = -\frac1n N \cdot N^*
\vspace{-2mm}
\end{equation}
where $N$ and $N^*$ are predicted and ground truth normal vector maps, and
the sums again run over valid pixels (\ie those with a ground truth normal).

For ground truth targets, we compute the normal map using the same
method as in Silberman \etal \cite{Silberman12}, which estimates normals from 
depth by fitting least-squares planes to neighboring sets of points in the
point cloud.

\vspace{-1mm}
\subsection{Semantic Labels}
\label{sec:semantic}
\vspace{-2mm}

For semantic labeling, we use a pixelwise softmax classifier to predict a class
label for each pixel.  The final output then has as many channels as there are
classes. We use a simple pixelwise cross-entropy loss,
\vspace{-1mm}
\begin{equation}
L_{semantic}(C, C^*) = -\frac1n \sum_i C_i^* \log(C_i)
\vspace{-2mm}
\end{equation}
where $C_i = e^{z_i} / \sum_c e^{z_{i,c}}$ is the class prediction
at pixel $i$ given the output $z$ of the final convolutional linear layer 3.4.

When labeling the NYUDepth RGB-D dataset, we use the ground truth depth and normals as
additional input channels.  We convolve each of the three input types (RGB,
depth and normals) with a different set of $32\times9\times9$ filters, then
concatenate the resulting three feature sets along with the network output from
the previous scale to form the input to the next.  We also tried the ``HHA'' encoding proposed by
\cite{gupta14}, but did not see a benefit in our case, thus we opt for the simpler
approach of using the depth and $xyz$-normals directly.  Note the first scale
is initialized using ImageNet, and we keep it RGB-only.  Applying convolutions
to each input type separately, rather than concatenating all the channels
together in pixel space and filtering the joint input, enforces independence
between the features at the lowest filter level, which we found helped
performance.

\section{Training}
\vspace{-2mm}

\subsection{Training Procedure}
\vspace{-2mm}

We train our model in two phases using SGD:  First, we jointly train both
Scales 1 and 2.  Second, we fix the parameters of these scales and train Scale
3.
Since Scale 3 contains four times as many pixels as Scale 2, it is
expensive to train using the entire image area for each gradient step.  To
speed up training, we instead use random crops of size
74x55:  We first forward-propagate the entire image through scales 1 and 2,
upsample, and crop the resulting Scale 3 input, as well as the original
RGB input at the corresponding location.  The cropped image and Scale 2
prediction are forward- and back-propagated through the Scale 3 network, and
the weights updated.  We find this speeds up
training by about a factor of 3, including the overhead for inference of the
first two scales, and results in about the same if not slightly better error
from the increased stochasticity.

All three tasks use the same initialization and learning rates in nearly all
layers, indicating that hyperparameter settings are in fact fairly robust to
changes in task.  Each were first tuned using the depth task, then verified to
be an appropriate order of magnitude for each other task using a small
validation set of 50 scenes.  The only differences are:
\enum{i} The learning rate for the normals task is 10 times larger than depth or labels.
\enum{ii} Relative learning rates of layers 1.6 and 1.7 are 0.1 each for
depth/normals, but 1.0 and 0.01 for semantic labeling.
\enum{iii} The dropout rate of layer 1.6 is 0.5 for depth/normals, but 0.8 for semantic labels, as there are fewer training images.

We initialize the convolutional layers in Scale 1 using ImageNet-trained
weights, and randomly initialize the
fully connected layers of Scale 1 and all layers in Scales 2 and 3.  We train
using batches of size 32 for the AlexNet-initialized model but batches of size
16 for the VGG-initialized model due to memory constraints.  In each case we
step down the global learning rate by a factor of 10 after approximately 2M
gradient steps, and train for an additional 0.5M steps.

\vspace{-1mm}
\subsection{Data Augmentation}
\vspace{-2mm}

In all cases, we apply random data transforms to augment the training data.  We
use random scaling, in-plane rotation, translation, color, flips and contrast.
When transforming an input and target, we apply corresponding transformations
to RGB, depth, normals and labels.  Note the normal vector transformation is
the inverse-transpose of the worldspace transform:  Flips and in-plane
rotations require flipping or rotating the normals, while to scale the image by
a factor $s$, we divide the depths by $s$ but multiply the $z$ coordinate of
the normals and renormalize.

\vspace{-1mm}
\subsection{Combining Depth and Normals}
\vspace{-2mm}

We combine both depths and normals networks together to share computation,
creating a network using a single scale 1 stack, but separate scale 2 and 3
stacks.  Thus we predict both depth and normals at the same time, given an RGB
image. This produces a 1.6x speedup compared to using two separate models.
\footnote{This shared model also enabled us to try enforcing compatibility
between predicted normals and those obtained via finite difference of the
predicted depth (predicting normals directly performs considerably better than
using finite difference).  However, while this constraint was able to improve
the normals from finite difference, it failed to improve either task
individually. Thus, while we make use of the shared model for computational
efficiency, we do not use the extra compatibility constraint.}

\vspace{-3mm}
\section{Performance Experiments}
\vspace{-2mm}

\subsection{Depth}
\vspace{-2mm}

We first apply our method to depth prediction on NYUDepth v2.  
We train using the entire NYUDepth v2 raw data distribution,
using the scene split specified in the official train/test distribution.  We
then test on the common distribution depth maps, including filled-in areas, but
constrained to the axis-aligned rectangle where there there is a valid depth
map projection.  Since the network output is a lower resolution than the
original NYUDepth images, and excludes a small border, we bilinearly upsample
our network outputs to the original 640x480 image scale, and extrapolate the
missing border using a cross-bilateral filter.
We compare our
method to prior works Ladicky \etal \cite{ladicky14depth}, Karsh \etal
\cite{Karsch14}, Baig \etal \cite{baig15}, Liu \etal \cite{liu14} and Eigen \etal \cite{depth}.

{
\begin{figure}[t]
\centering
\includegraphics[width=\columnwidth]{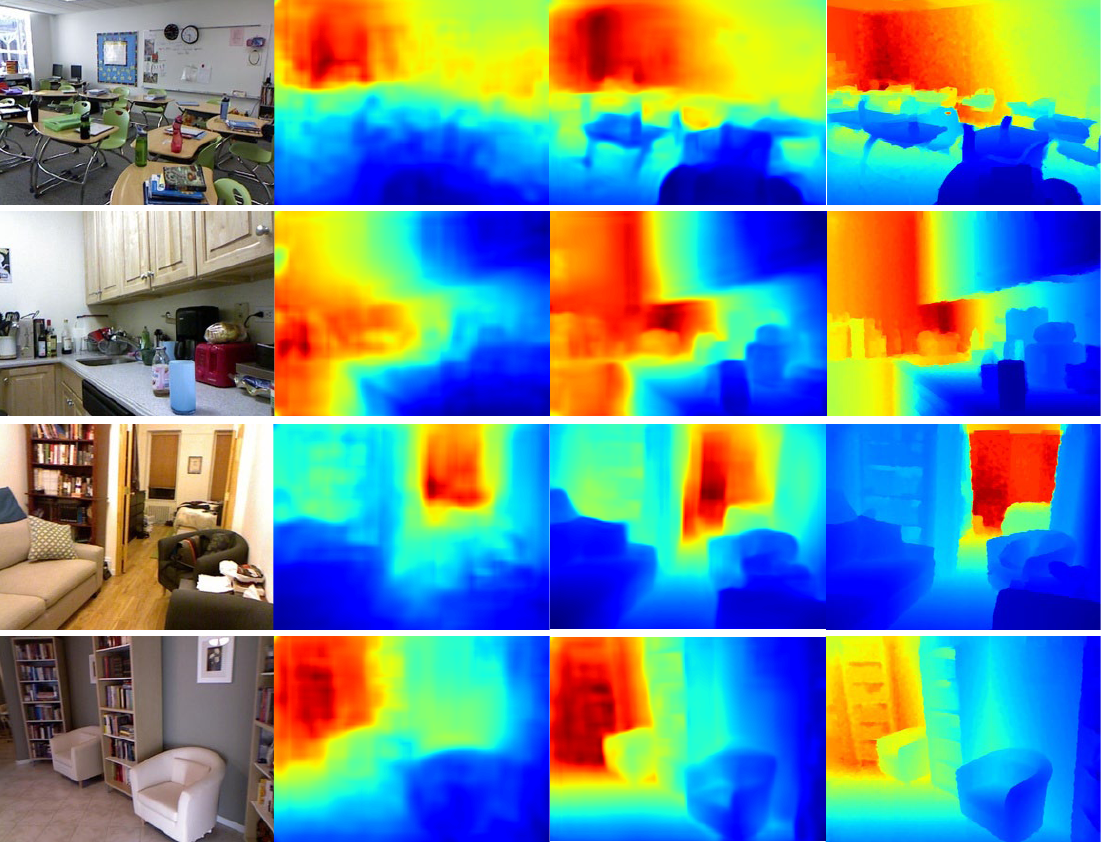}
\scriptsize
\setlength{\tabcolsep}{25pt}
\begin{tabular}{cccc}
(a)&(b)&(c)&(d)
\end{tabular}
\caption{Example depth results. (a) RGB input; (b) result of \cite{depth};
(c) our result; (d) ground truth.  Note the color range of each image
is individually scaled.
}
\label{fig:depth-examples}
\vspace{-2mm}
\end{figure}
}

\begin{table}[t]
\centering
\scriptsize

\setlength{\tabcolsep}{2pt}
\begin{tabular}{|l||ccccccc|}
\hline
\multicolumn{8}{|c|}{\bf \small Depth Prediction} \\
\hline
\hline
 & \hspace{-2mm}Ladicky\cite{ladicky14depth} & \hspace{-2mm}Karsch\cite{Karsch14} & Baig \cite{baig15} & Liu \cite{liu14} & \hspace{-1mm} Eigen\cite{depth} & Ours(A) & Ours(VGG) \\
\hline                           
\hline
          $\delta<1.25$         &   0.542  &  --     & 0.597  &  0.614   &  0.614    &   0.697  &  {\bf 0.769}  \\ 
          $\delta<1.25^2$       &   0.829  &  --     & --     &  0.883   &  0.888    &   0.912  &  {\bf 0.950}  \\ 
          $\delta<1.25^3$       &   0.940  &  --     & --     &  0.971   &  0.972    &   0.977  &  {\bf 0.988}  \\ 
\hline                                                                                          
abs rel                           &    --    &  0.350  & 0.259  &  0.230   &  0.214    &   0.198  &  {\bf 0.158}  \\ 
sqr rel                           &    --    &  --     & --     &   --     &  0.204    &   0.180  &  {\bf 0.121}  \\ 
 RMS(lin)                    &    --    &  1.2    & 0.839  &  0.824   &  0.877    &   0.753  &  {\bf 0.641}  \\ 
 RMS(log)                       &    --    &  --     & --     &   --     &  0.283    &   0.255  &  {\bf 0.214}  \\ 
 sc-inv.            &    --    &  --     & 0.242  &   --     &  0.219    &   0.202  &  {\bf 0.171}  \\ 
\hline
\end{tabular}

\caption{Depth estimation measurements.  Note higher is better for top rows
of the table, while lower is better for the bottom section.}
\label{tab:depth}
\vspace{-4mm}
\end{table}

The results are shown in \tab{depth}.  Our model obtains best performance in
all metrics, due to our larger architecture and improved training.  In
addition, the VGG version of our model significantly outperforms the smaller
AlexNet version, reenforcing the importance of model size; this is the case
even though the depth task is seemingly far removed from the classification
task with which the initial coarse weights were first trained.
Qualitative results in
\fig{depth-examples} show substantial improvement in detail sharpness over
\cite{depth}.

\vspace{-1mm}
\subsection{Surface Normals}
\vspace{-3mm}

Next we apply our method to surface normals prediction.  We compare against the
3D Primitives (3DP) and ``Indoor Origami'' works of Fouhey \etal
\cite{fouhey13,fouhey14}, Ladicky \etal \cite{ladicky14normals}, and Wang \etal
\cite{wang14}.  As with the depth network, we used the full raw dataset for
training, since ground-truth normal maps can be generated for all images.
Since different systems have different ways of calculating ground truth normal
maps, we compare using both the ground truth as constructed in
\cite{ladicky14normals} as well as the method used in \cite{Silberman12}.
The differences between ground
truths are due primarily to the fact that \cite{ladicky14normals} uses more aggressive smoothing; 
thus \cite{ladicky14normals} tends to present flatter areas,
while \cite{Silberman12} is noisier but keeps more details present.
We measure performance with the same metrics as in \cite{fouhey13}:  The mean
and median angle from the ground truth across all unmasked pixels, as well as
the percent of vectors whose angle falls within three thresholds.

Results are shown in \tab{normals}.  The smaller version of our model performs
similarly or slightly better than Wang {\it et al.}, while the larger version
substantially outperforms all comparison methods.
Figure \ref{fig:norm-compare} shows example predictions.  Note the details
captured by our method, such as the curvature of the blanket on
the bed in the first row, sofas in the second row, and objects in the last
row.

\begin{table}[t]
\centering
\small

\scriptsize

\begin{tabular}{|l||cc|ccc|}
\hline
\multicolumn{6}{|c|}{\bf \small Surface Normal Estimation (GT \cite{ladicky14normals})} \\
\hline
\hline
& \multicolumn{2}{c|}{\bf Angle Distance} 
& \multicolumn{3}{c|}{\bf Within $t^\circ$ Deg.} \\
& {\bf Mean} & {\bf Median} 
& $11.25^\circ$ & $22.5^\circ$ & $30^\circ$ \\
\hline
\hline
3DP          \cite{fouhey13}          &    35.3     &    31.2     &      16.4        &       36.6      &       48.2       \\
Ladicky \aal \cite{ladicky14normals}  &    33.5     &    23.1     &      27.5        &       49.0      &       58.7       \\
Fouhey \aal  \cite{fouhey14}          &    35.2     &    17.9     &      40.5        &       54.1      &       58.9       \\
Wang \aal    \cite{wang14}            &    26.9     &    14.8     &      42.0        &       61.2      &       68.2       \\
\hline
Ours (AlexNet)                        &    23.7     &    15.5     &      39.2        &       62.0      &       71.1      \\
Ours (VGG)                            & {\bf 20.9}  & {\bf 13.2}  & {\bf 44.4}       & {\bf  67.2}     & {\bf  75.9}     \\
\hline
\end{tabular}

\begin{tabular}{|l||cc|ccc|}
\hline
\multicolumn{6}{|c|}{\bf \small Surface Normal Estimation (GT \cite{Silberman12})} \\
\hline
\hline
& \multicolumn{2}{c|}{\bf Angle Distance} 
& \multicolumn{3}{c|}{\bf Within $t^\circ$ Deg.} \\
& {\bf Mean} & {\bf Median} 
& $11.25^\circ$ & $22.5^\circ$ & $30^\circ$ \\
\hline
\hline
3DP          \cite{fouhey13}          &   37.7      &   34.1       &   14.0      &   32.7     &    44.1     \\
Ladicky \aal \cite{ladicky14normals}  &   35.5      &   25.5       &   24.0      &   45.6     &    55.9     \\
Wang \aal    \cite{wang14}            &   28.8      &   17.9       &   35.2      &   57.1     &    65.5     \\
\hline
Ours (AlexNet)                        &   25.9      &   18.2       &   33.2      &   57.5     &    67.7     \\
Ours (VGG)                            & {\bf 22.2}  & {\bf  15.3}  & {\bf 38.6}  & {\bf 64.0} & {\bf 73.9}  \\
\hline
\end{tabular}

\caption{Surface normals prediction measured against the ground truth constructed by \cite{ladicky14normals} (top) and \cite{Silberman12} (bottom).}
\label{tab:normals}
\vspace{-4mm}
\end{table}

\vspace{-1mm}
\subsection{Semantic Labels}
\vspace{-1mm}

\subsubsection{NYU Depth}
\vspace{-2mm}

We finally apply our method to semantic segmentation, first also on NYUDepth.
Because this data provides a depth channel, we use the ground-truth depth and normals as
input into the semantic segmentation network, as described in \secc{semantic}.
We evaluate our method on semantic class sets with 4, 13 and 40 labels,
described in \cite{Silberman12}, \cite{couprie13} and \cite{gupta13},
respectively.  The 4-class segmentation task uses high-level category labels
``floor'', ``structure'', ``furniture'' and ``props'', while the 13- and
40-class tasks use different sets of more fine-grained categories.  We compare
with several recent methods, using the metrics commonly used to evaluate
each task:  For the 4- and 13-class tasks we use pixelwise and per-class
accuracy; for the 40-class task, we also compare using the mean pixel-frequency
weighted Jaccard index of each class, and the flat mean Jaccard index.

Results are shown in \tab{nyud40}.  We decisively outperform the comparison
methods on the 4- and 14-class tasks.  In the 40-class task, our model
outperforms Gupta \etal '14 with both model sizes, and Long \etal with the
larger size.  Qualitative results are shown in \fig{sem-examples-nyud}.  Even
though our method does not use superpixels or any piecewise constant
assumptions, it nevertheless tends to produce large constant regions most of
the time.

\begin{table}[t]
\centering
\small
\begin{tabular}{cc}
\hspace{-3mm}
\scriptsize
\begin{tabular}{|l||cc|}
\hline
\multicolumn{3}{|c|}{\bf \small 4-Class Semantic Segmentation} \\
\hline
              & Pixel & Class \\
\hline
\hline
Couprie \aal  \cite{couprie13} &   64.5    &    63.5        \\
Khan    \aal  \cite{khan14}    &  69.2     &    65.6        \\
Stuckler \aal \cite{stuckler} &  70.9     &   67.0         \\  
Mueller \aal \cite{mueller}   &  72.3     &   71.9         \\  
Gupta \aal '13 \cite{gupta13}  & 78   &  -- \\ 
\hline
Ours (AlexNet) &         80.6    &         79.1       \\  
Ours (VGG)     &    {\bf 83.2}   &    {\bf 82.0}       \\  
\hline
\end{tabular}

&
\hspace{-5mm}
\scriptsize
\begin{tabular}{|l||cc|}
\hline
\multicolumn{3}{|c|}{\bf \small 13-Class Semantic} \\
\hline
              & Pixel & Class \\
\hline
\hline
Couprie \aal \cite{couprie13} &  52.4     &   36.2         \\
Wang    \aal \cite{wang14label} &   --      &   42.2         \\
Hermans \aal \cite{hermans14} &  54.2     &   48.0         \\
Khan \aal \cite{khan14} $^*$ &  58.3   &   45.1         \\
\hline
Ours (AlexNet) &       70.5      &        59.4          \\  
Ours (VGG)     &  {\bf 75.4}     &   {\bf 66.9}         \\  
\hline
\multicolumn{3}{c}{} \\
\end{tabular}

\end{tabular}
\scriptsize
\begin{tabular}{|l||cccc|}
\hline
\multicolumn{5}{|c|}{\bf \small 40-Class Semantic Segmentation} \\
\hline
& Pix. Acc. & Per-Cls Acc. & Freq. Jaccard & Av. Jaccard \\
\hline
\hline
Gupta{\it\&al.}'13 \cite{gupta13} &   59.1 & 28.4 & 45.6 & 27.4 \\
Gupta{\it\&al.}'14 \cite{gupta14} &   60.3 & 35.1 & 47.0 & 28.6  \\
Long{\it\&al.} \cite{long14}      &   65.4 & {\bf 46.1} & 49.5 & 34.0 \\
\hline
Ours (AlexNet)    &        62.9  &      41.3  &      47.6  &      30.8  \\
Ours (VGG)        &   {\bf 65.6} &      45.1  & {\bf 51.4} & {\bf 34.1} \\
\hline
\end{tabular}

\caption{Semantic labeling on NYUDepth v2}
{\scriptsize $^*$Khan\aal use a different overlapping label set.}
\label{tab:nyud4}
\label{tab:nyud14}
\label{tab:nyud40}
\vspace{-2mm}
\end{table}

\vspace{-7mm}
\subsubsection{Sift Flow}
\vspace{-2mm}

We confirm our method can be applied to additional scene types by
evaluating on the Sift Flow dataset \cite{siftflow}, which
contains images of outdoor cityscapes and landscapes segmented into 33
categories.  We found no need to adjust
convolutional kernel sizes or learning rates for this dataset, and
simply transfer the values used for NYUDepth directly; however, 
we do adjust the output sizes of the layers to match the new image sizes.

We compare against Tighe \etal \cite{tighe2013finding}, Farabet \etal
\cite{farabet2012scene}, Pinheiro \cite{pinheiro14} and Long \etal
\cite{long14}.  Note that Farabet \etal train two models, using 
empirical or rebalanced class distributions by resampling superpixels.
We train a more class-balanced version of our model by reweighting
each class in the cross-entropy loss;  we weight each pixel by 
$\alpha_c = median\_freq / freq(c)$ where $freq(c)$ is the number of pixels of
class $c$ divided by the total number of pixels in images where $c$ is present,
and $median\_freq$ is the median of these frequencies.

Results are in \tab{siftflow}; we compare regular (1) and reweighted (2)
versions of our model against comparison methods.  Our smaller model 
substantially outperforms all
but Long \etal, while our larger model performs
similarly to Long \etal  This
demonstrates our model's adaptability not just to different tasks but also
different data.

\begin{table}[t]
\centering
\scriptsize
\begin{tabular}{|l||cccc|}
\hline
\multicolumn{5}{|c|}{\bf \small Sift Flow Semantic Segmentation} \\
\hline
\hline
              & Pix. Acc. & Per-Class Acc. & Freq. Jacc & Av. Jacc \\
\hline
\hline
Farabet \aal (1) \cite{farabet2012scene} & 78.5    &  29.6  & -- & -- \\ 
Farabet \aal (2) \cite{farabet2012scene} & 74.2    &  46.0  & -- & -- \\
Tighe \aal \cite{tighe2013finding}       &  78.6   &  39.2  & -- & -- \\
Pinheiro \aal \cite{pinheiro14}          &  77.7   &  29.8  & -- & -- \\
Long \aal \cite{long14}                  &  85.1   &  51.7  & 76.1 & {\bf 39.5}\\
\hline
Ours (AlexNet) (1)                       &  84.0   &  42.0  & 73.7 & 33.1    \\
Ours (AlexNet) (2)                       &  81.6   &  48.2  & 71.3 & 32.6    \\
Ours (VGG) (1)                           & {\bf 86.8}  & 46.4  & {\bf 77.9} & 38.8 \\
Ours (VGG) (2)                           &  83.8   & {\bf 55.7} & 74.7 & 37.6 \\
\hline
\end{tabular}
\caption{Semantic labeling on the Sift Flow dataset. (1) and (2) correspond to
non-reweighted and class-reweighted versions of our model (see text).}
\label{tab:siftflow}
\vspace{-3mm}
\end{table}

\begin{table}[h]
\centering
\scriptsize

{
\setlength{\tabcolsep}{1pt}
\begin{tabular}{|l||cccc||c||c|}
\hline
\multicolumn{7}{|c|}{\bf \small Pascal VOC Semantic Segmentation} \\
\hline
&
\multicolumn{4}{c||}{\small 2011 Validation}
&
\multicolumn{1}{c||}{\small 2011 Test} 
&
\multicolumn{1}{c|}{\small 2012 Test} \\
\hline
& Pix. Acc. & Per-Cls Acc. & Freq.Jacc & Av.Jacc & Av.Jacc & Av.Jacc \\
\hline
\hline
Dai{\it\&al.}\cite{dai14}        &  --         &  --        &   --       &   --       &   --       &  61.8 \\
Long{\it\&al.}\cite{long14}      &  90.3 & 75.9 & 83.2 & 62.7 & {\bf 62.7} & 62.2 \\
Chen{\it\&al.}\cite{chen15}      &  --         &  --        &   --       &   --       &   --       & {\bf 71.6} \\
\hline
Ours (VGG)        &   90.3 & 72.4 & 82.9 & 62.2 & 62.5 & 62.6 \\

\hline
\end{tabular}
}
\vspace{-2mm}
\caption{Semantic labeling on Pascal VOC 2011 and 2012.}
\label{tab:voc}
\vspace{-3mm}
\end{table}

\vspace{-4mm}
\subsubsection{Pascal VOC}
\vspace{-2mm}

In addition, we also verify our method using Pascal VOC.
Similarly to Long \etal \cite{long14}, we train using the 2011 training
set augmented with 8498 training images collected by Hariharan \etal
\cite{hariharan2014}, and evaluate using the 736 images from the 2011
validation set not also in the Hariharan extra set, as well as on the
2011 and 2012 test sets.
We perform online data
augmentations as in our NYUDepth and Sift Flow models, and use the same
learning rates. Because these images
have arbitrary aspect ratio, we train our model on
square inputs, and scale the smaller side of each image to 256;
at test time we apply the model with a stride of 128 to cover the image
(two applications are usually sufficient).

Results are shown in \tab{voc} and \fig{sem-examples-voc}.  We compare
with Dai \etal \cite{dai14}, Long \etal\cite{long14} and Chen \etal\cite{chen15}; the latter is a more recent work
that augments a convolutional network with large top-layer field of and fully-connected CRF.  Our model
performs comparably to Long {\it et al.}, even as it generalizes to multiple
tasks, demonstrated by its adeptness at depth and normals prediction.

\vspace{-2mm}
\section{Probe Experiments}
\vspace{-1mm}

\subsection{Contributions of Scales}
\vspace{-2mm}
\label{sec:scales}

We compare performance broken down according to the different
scales in our model in \tab{nyudscale}.  For depth, normals and 4- and 13-class
semantic labeling tasks, we train and evaluate the model using just scale 1,
just scale 2, both, or all three scales 1, 2 and 3.  For the coarse scale-1-only
prediction, we replace the last fully connected layer of the coarse stack with a
fully connected layer that outputs directly to target size, \ie a pixel map of
either 1, 3, 4 or 13 channels depending on the task.  The spatial resolution is
the same as is used for the coarse features in our model, and is upsampled in
the same way.

We report the ``abs relative difference'' measure (\ie $|D-D^*|/D^*$)
to compare depth, mean
angle distance for normals, and pixelwise accuracy for semantic segmentation.

\begin{table}[t]
\centering
\scriptsize
\setlength{\tabcolsep}{4pt}
\begin{tabular}{|l||c|c|cc|cc|}
\hline
\multicolumn{7}{|c|}{\bf \small Contributions of Scales} \\
\hline
\hline
& \multirow{2}{*}{\bf Depth}
& \multirow{2}{*}{\bf Normals}
& \multicolumn{2}{c|}{\bf 4-Class}
& \multicolumn{2}{c|}{\bf 13-Class} \\
& &
& RGB+D+N & RGB 
& RGB+D+N & RGB \\
\hline
& \multicolumn{2}{c|}{Pixelwise Error}
& \multicolumn{4}{c|}{Pixelwise Accuracy} \\
& \multicolumn{2}{c|}{lower is better}
& \multicolumn{4}{c|}{higher is better} \\
\hline
\hline
Scale 1 only  
    & \underline{0.218} & \underline{29.7}
    &   71.5    &   \underline{71.5}  
    &    58.1   &  \underline{58.1} \\
Scale 2 only
    &  0.290  &   31.8  
    & \underline{77.4} & 67.2
    & \underline{65.1} &   53.1   \\
Scales 1 + 2 
    &  0.216  &   26.1    &   80.1  & 74.4  &    69.8   & 63.2 \\
Scales 1 + 2 + 3                          
    &  0.198  &   25.9    &   80.6  & 75.3  &    70.5   & 64.0 \\
\hline
\end{tabular}
\caption{Comparison of networks for different scales for depth, normals and semantic labeling tasks with 4 and 13 categories.  Largest single contributing scale
is underlined.}
\vspace{-3mm}
\label{tab:nyudscale}
\end{table}

\begin{table}
\centering
\scriptsize
\begin{tabular}{|l||cc|cc|}
\hline
\multicolumn{5}{|c|}{\bf \small Effect of Depth/Normals Inputs} \\
\hline
\hline
              & \multicolumn{2}{c|}{Scale 2 only}
              & \multicolumn{2}{c|}{Scales 1 + 2} \\
              & Pix. Acc. & Per-class
              & Pix. Acc. & Per-class \\
\hline
\hline
RGB only           & 53.1 & 38.3 & 63.2 & 50.6 \\
RGB + pred. D\&N   & 58.7 & 43.8 & 65.0 & 49.5 \\
RGB + g.t. D\&N    & 65.1 & 52.3 & 69.8 & 58.9 \\
\hline
\end{tabular}
\caption{Comparison of RGB-only, predicted depth/normals, and ground-truth depth/normals as input to the
13-class semantic task.
}
\label{tab:nyudpredlabels}
\vspace{-4mm}
\end{table}

{
\begin{figure*}[!htbp]
\centering
\def\mycolwd{70pt}
\begin{tabular}{p{\mycolwd}p{\mycolwd}p{\mycolwd}p{\mycolwd}p{\mycolwd}p{\mycolwd}}
 \centering \small \bf RGB input 
&\centering \small \bf 3DP \cite{fouhey13}
&\centering \small \bf Ladicky\aal \cite{ladicky14normals}
&\centering \small \bf Wang\aal \cite{wang14}
&\centering \small \bf Ours (VGG)
&\centering \small \bf Ground Truth
\end{tabular}
\includegraphics[width=\textwidth]{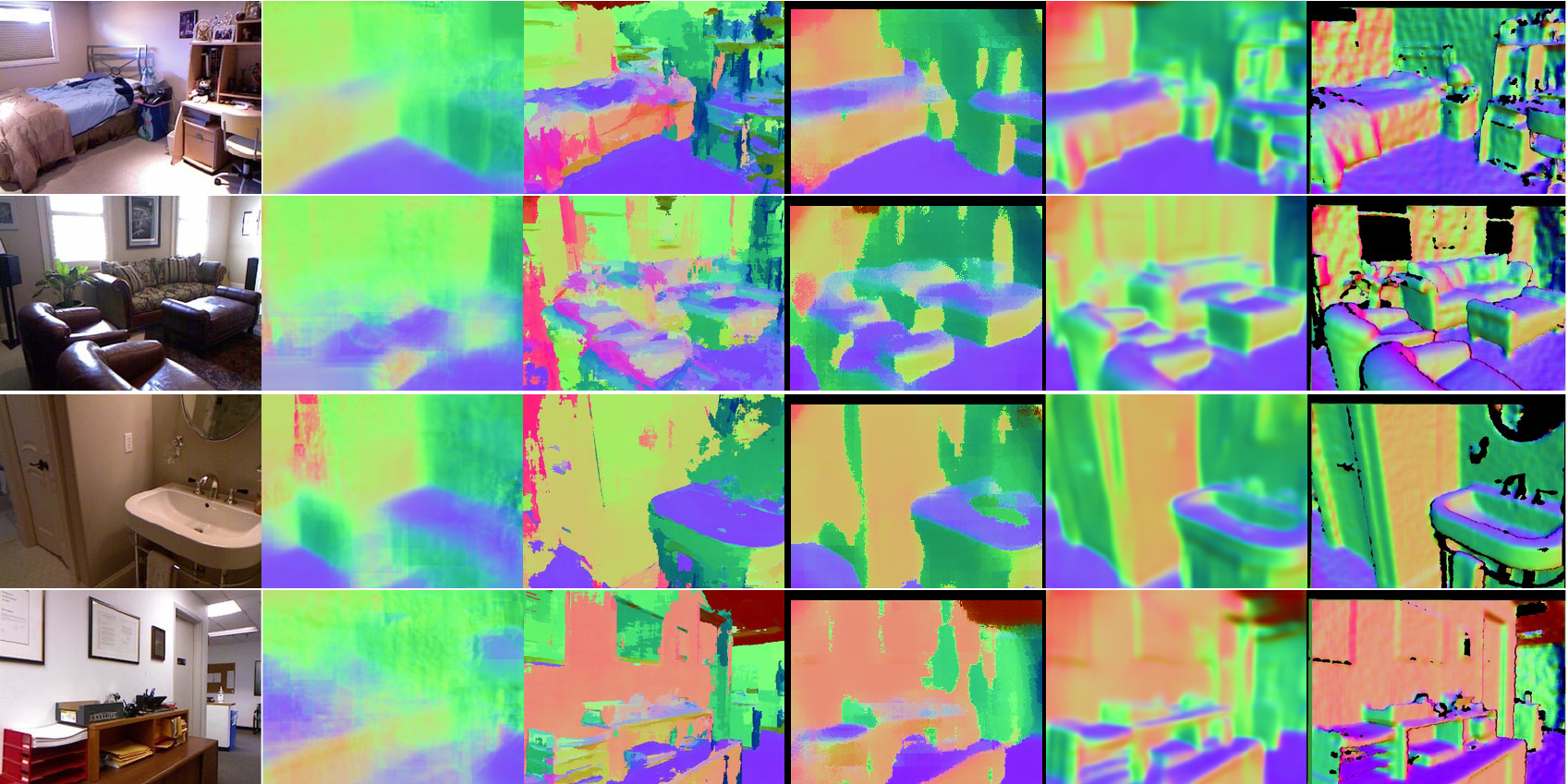}
\caption{Comparison of surface normal maps.}
\label{fig:norm-compare}
\vspace{-4mm}
\end{figure*}
}

First, we note there is progressive improvement in all tasks as scales are
added (rows 1, 3, and 4).  In addition, we find the largest single contribution
to performance is the coarse Scale 1 for depth and normals, but the more local
Scale 2 for the semantic tasks --- however, this is only due to the fact that
the depth and normals channels are introduced at Scale 2 for the semantic
labeling task.  Looking at the labeling network with RGB-only inputs, we find
that the coarse scale is again the larger contributer,
indicating the
importance of the global view.
(Of course, this scale was
also initialized with ImageNet convolution weights that are much related to
the semantic task; however, even initializing randomly achieves 
54.5\% for 13-class scale 1 only, still the largest contribution, albeit by a
smaller amount).

\vspace{-1mm}
\subsection{Effect of Depth and Normals Inputs}
\vspace{-2mm}

The fact that we can recover much of the depth and normals information from
the RGB image naturally leads to two questions:  \enum{i} How important
are the depth and normals inputs relative to RGB in the semantic
labeling task?  \enum{ii} What might happen if we were to replace the true
depth and normals inputs with the predictions made by our network?

To study this, we trained and tested our network using either Scale 2 alone or
both Scales 1 and 2 for the 13-class semantic labeling task under three input
conditions: \enum{a} the RGB image only, \enum{b} the RGB image along with
predicted depth and normals, or \enum{c} RGB plus true depth
and normals.  Results are in \tab{nyudpredlabels}.  Using ground truth
depth/normals shows substantial improvements over RGB alone.  Predicted
depth/normals appear to have little effect when using both scales, but
a tangible improvement when using only Scale 2.  We believe this is because
any relevant information provided by predicted depths/normals for labeling can
also be extracted from the input; thus the labeling network can learn this same
information itself, just from the label targets.  However, this supposes that the
network structure is capable of learning these relations:  If this is not the
case, \eg when using only Scale 2, we do see improvement.  This is also
consistent with \secc{scales}, where we found the coarse network was important
for prediction in all tasks --- indeed, supplying the predicted
depth/normals to scale 2 is able to recover much of the performance
obtained by the RGB-only scales 1+2 model.

\vspace{-2mm}
\section{Discussion}
\vspace{-2mm}

Together, depth, surface normals and semantic labels provide a rich
account of a scene. We have proposed a simple and fast multiscale
architecture using convolutional networks that gives excellent 
performance on all three modalities. The models beat existing methods
on the vast majority of benchmarks we explored. This is impressive
given that many of these methods are specific to a single modality and
often slower and more complex algorithms than ours. As such, our model
provides a convenient new baseline for the three tasks. To this end, code and trained
models can be found at \url{http://cs.nyu.edu/~deigen/dnl/}. 

\clearpage

{\onecolumn
\begin{figure*}
\centering
\def\mycolwd{110pt}
\begin{tabular}{p{\mycolwd}p{\mycolwd}p{\mycolwd}p{\mycolwd}}
 \centering \small \bf RGB input 
&\centering \small \bf 4-Class Prediction
&\centering \small \bf ~~13-Class Prediction
&\centering \small \bf \quad13-Class Ground Truth
\end{tabular}
\includegraphics[width=\textwidth]{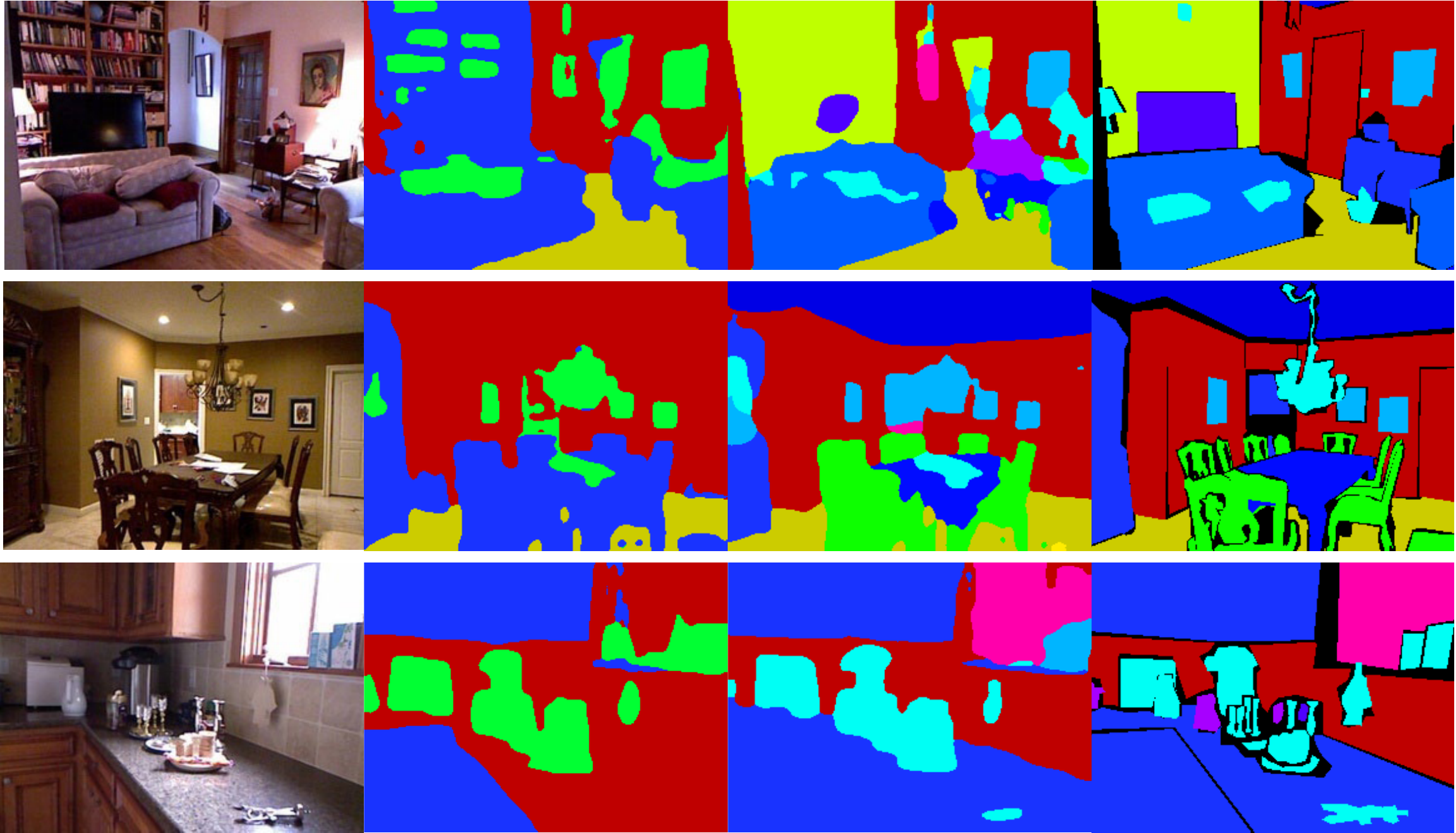}
\vspace{-2mm}
\caption{Example semantic labeling results for NYUDepth:
(a) input image; (b) 4-class labeling result; (c)
13-class result; (d) 13-class ground truth.
} \label{fig:sem-examples-nyud}
\end{figure*}
\vspace{-5mm}
\begin{figure*}
\centering
\includegraphics[width=\textwidth]{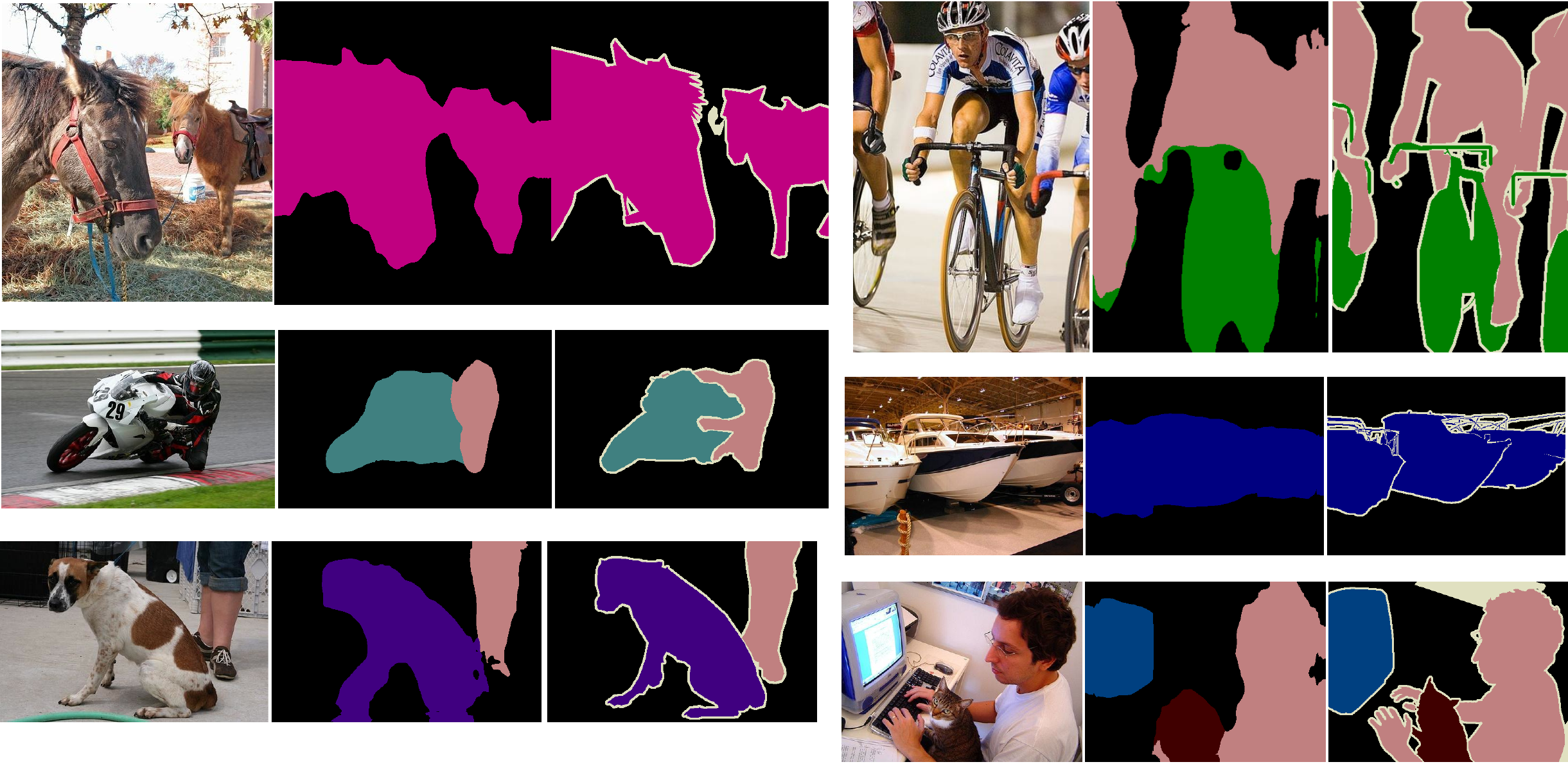}
\caption{Example semantic labeling results for Pascal VOC 2011.  For
each image, we show RGB input, our prediction, and ground truth.
} \label{fig:sem-examples-voc}
\end{figure*}

\section*{Acknowledgements}
\vspace{-2mm}
This work was supported by an ONR \#N00014-13-1-0646 and an NSF CAREER grant.
\vspace{-2mm}
}
\twocolumn

\clearpage

{
\small
\bibliographystyle{ieee}
\bibliography{iccv15}
}

\end{document}